%% file: _publication_acl.tex
\documentclass{article}
\usepackage[preprint]{neurips_2023}
\input{common}

\title{Towards Improving Robustness Against \\Common Corruptions in Object Detectors \\Using Adversarial Contrastive Learning}
\author{%
  Shashank Kotyan \quad Danilo Vasconcellos Vargas \\
  Laboratory of Intelligent Systems \\
  Kyushu Univeristy, Fukuoka, Japan \\
}

\begin{document}

\maketitle

\begin{abstract}
Neural networks have revolutionized various domains, exhibiting remarkable accuracy in tasks like natural language processing and computer vision. 
However, their vulnerability to slight alterations in input samples poses challenges, particularly in safety-critical applications like autonomous driving. 
Current approaches, such as introducing distortions during training, fall short in addressing unforeseen corruptions. 
This paper proposes an innovative adversarial contrastive learning framework to enhance neural network robustness simultaneously against adversarial attacks and common corruptions. 
By generating instance-wise adversarial examples and optimizing contrastive loss, our method fosters representations that resist adversarial perturbations and remain robust in real-world scenarios.
Subsequent contrastive learning then strengthens the similarity between clean samples and their adversarial counterparts, fostering representations resistant to both adversarial attacks and common distortions. 
By focusing on improving performance under adversarial and real-world conditions, our approach aims to bolster the robustness of neural networks in safety-critical applications, such as autonomous vehicles navigating unpredictable weather conditions. 
We anticipate that this framework will contribute to advancing the reliability of neural networks in challenging environments, facilitating their widespread adoption in mission-critical scenarios.
\end{abstract}

\section{Introduction}

We've harnessed the power of neural networks to achieve high accuracy in a range of tasks in applied in different domains like natural language processing and computer vision.
Many of these tasks would be impractical without the support of neural networks.
However, current neural networks~\citep{krizhevsky2012ImageNet} don't exhibit the robustness seen in the human visual system~\citep{recht2018CIFAR10, recht2019ImageNet, azulay2019Why}.
Neural networks tend to make errors when faced with slight alterations in input samples~\citep{szegedy2014Intriguing, carlini2016Defensive, carlini2017Adversarial}, a contrast to the resilience of the human vision system against minor image changes~\citep{dodge2017Study}.
Humans effortlessly navigate through various corruptions and distortions in image, and even abstract changes in structure and style don't throw them off.

This susceptibility of neural networks poses a significant challenge in deploying them for safety-critical applications like autonomous driving.
Interestingly, a significant factor hindering the mainstream adoption of autonomous cars is the inadequacy of their recognition models in adverse weather conditions, as noted by~\cite{dai2018Dark, michaelis2020Benchmarking}.
Such failure is detrimental for mission-critical applications such as self-driving, in which domain shifts are inevitable, and it is important for models to be robust to domain shifts to make them reliable in real-world conditions.

Previous studies propose enhancing the training data by introducing various distortions to enhance overall robustness.
Acquiring data for unusual weather conditions is challenging, and while various common environmental conditions, such as fog~\citep{sakaridis2018Semantic}, rain~\citep{bernuth2019Simulating}, snow~\citep{bernuth2019Simulating}, and daytime to nighttime transitions~\citep{dai2018Dark}, have been modeled, it remains impossible to anticipate all potential conditions that might arise in a real-world situation.
Further, this approach falls short when dealing with unforeseen corruptions.
Recent findings indicate that neural networks exhibit poor generalization to newly introduced distortion types, even when trained on a diverse set of other distortions, as illustrated by~\cite{geirhos2018Generalisation}.
In a broader context, neural networks frequently struggle to extend their performance beyond the domain or distribution of the training data.
Consequently, the generation of adversarial examples, as demonstrated by~\cite{szegedy2014Intriguing}, becomes feasible.

% Examples include the failure to generalize to images with uncommon poses of objects [Alcorn et al., 2019] or to cope with small distributional changes [e.g. Zech et al., 2018, Touvron et al., 2019].

Adversarial examples represent the most extreme attempts to exploit the weaknesses of neural networks which characterized by a minimal domain shift, imperceptible to humans yet sufficient to deceive a neural network.
Thus, analyzing with adversarial examples entails considering the worst-case scenario for a neural network rather than the more common average-case scenarios encountered in real-world situations.
To address the prevalent but less extreme issue of perceptible image distortions, such as blurriness, noise, or natural distortions like snow, one can employ the analysis provided by~\cite{hendrycks2019Benchmarking} using their proposed common corruptions benchmarks.
This allows for an understanding of how performance degrades in situations closely resembling real-world conditions.

If we could construct models resilient to every conceivable image corruption, the impact of weather changes on autonomous cars would likely be negligible.
However, gauging the robustness of models requires the establishment of a measurable criterion.
This, however, proves to be a challenging task due to the multitude of possibilities, definitions, exceptions, and trade-offs involved.
Conducting a comprehensive assessment of models across all potential corruption types is impractical.
Hence, it is suggested to evaluate models on a diverse range of corruption types, like the one in common corruptions benchmark~\citep{hendrycks2019Benchmarking}, that were not included in the training data.
We assert that this approach serves as a valuable approximation for predicting performance under natural distortions like rain, snow, fog, or transitions between day and night.
Enhancements in general robustness are anticipated to yield learning systems capable of more adept reasoning over data, thereby achieving a heightened level of abstraction.

\textbf{Contributions.}
\begin{description}[style=sameline, leftmargin=*]

    \item[Adversarial Contrastive Learning:]
    We propose an adversarial contrastive learning framework to train a neural network with adversarial robustness and contrastive loss simultaneously to improve the performance on common corruptions.
    Our idea is to first deceive the model by generating instance-wise adversarial examples.
    Specifically, we introduce perturbations to augmentations of the samples to maximize their contrastive loss, causing confusion in the instance-level classifier about the identities of the perturbed samples. 
    Subsequently, we enhance the similarity between clean samples and their adversarial counterparts using contrastive learning.
    This process leads to representations that suppress distortions caused by adversarial perturbations, ultimately resulting in learned representations that are robust against adversarial attacks and common corruptions, likewise.

\end{description}

\section{Related Works}

\subsection{Challenges in the field of Adversarial Machine Learning}

\textbf{Adversarial Examples. }
An adversarial image refers to a clean image that undergoes slight, carefully crafted distortions aimed at confusing a classifier.
These manipulative distortions can occasionally deceive black-box classifiers, as demonstrated by~\cite{kurakin2017Adversarial}.
Algorithms have been developed to identify the smallest additive distortions in RGB space capable of confusing a classifier, as outlined by~\cite{carlini2018GroundTruth}.
The peculiar behavior of neural networks for nearly identical images was revealed in~\cite{szegedy2014Intriguing}.
Subsequently,~\cite{nguyen2015Deep} illustrated that neural networks exhibit high confidence when confronted with textures and random noise, uncovering vulnerabilities in neural networks exploited by adversarial attacks.
The feasibility of universal adversarial perturbations, capable of deceiving a neural network across most samples, was demonstrated by~\cite{moosavi-dezfooli2017Universal} .
The introduction of patches into an image was also shown to lead to misclassification by neural networks~\citep{brown2018Adversarial}
Furthermore, an extreme attack proved effective, demonstrating the ability to cause misclassification with a single-pixel change~\citep{su2019One, kotyan2022Adversariala}.

\cite{goodfellow2015Explaining} proposed the Fast Gradient Sign Method (FGSM), which perturbs a target sample to its gradient direction, to increase its loss, and also use the generated samples to train the model for improved robustness. 
Follow-up works~\citep{madry2018Deepa, moosavi-dezfooli2016DeepFool, kurakin2017Adversarial, carlini2017Evaluating} proposed iterative variants of the gradient attack with improved adversarial learning frameworks.
Many of these attacks can be translated into real-world threats by printing out adversarial samples~\citep{kurakin2017Adversarial}.
Additionally, carefully crafted glasses and even general 3D adversarial objects have been shown to misclassify~\citep{sharif2016Accessorize, athalye2018Synthesizing}.
\cite{gilmer2018Motivating} proposed modifying the adversarial robustness problem for increased real-world applicability.

\textbf{Robustness against Adversarial Examples.}
Various defensive and detection systems have been proposed to mitigate adversarial attack problems.
Despite these efforts, there are currently no consistent solutions or promising ones to counter adversarial attacks.
Defensive systems come in many variations, as analyzed in detail in previous studies~\citep{dziugaite2016study, hazan2016Perturbations, buckman2018Thermometer, das2017Keeping, guo2018Countering, ma2018Characterizing, song2018PixelDefend, xu2018Feature}, and their shortcomings are documented~\citep{athalye2018Obfuscated, uesato2018Adversarial,kotyan2022Transferability}.

Defensive distillation, where a smaller neural network distills the knowledge learned by the original network, was proposed as a defense~\citep{papernot2016distillation}, but it was found to lack robustness~\citep{carlini2016Defensive, carlini2017Evaluating}.
The technique proposed by~\cite{xu2018Feature} involves comparing the prediction of a classifier with the prediction of the same input, but `squeezed,' enabling the detection of adversarial samples with small perturbations.
Adversarial training, where adversarial samples are used to augment the training dataset, has been proposed to increase robustness~\citep{goodfellow2015Explaining, huang2016Learning, madry2018Deepa, zhang2019Theoretically, tramer2019Adversarial, madaan2020Adversarial}, but resulting networks remain vulnerable to attacks~\citep{athalye2018Obfuscated, tramer2018Ensemble, tramer2020Adaptive, kotyan2022Adversariala}.
However, many detection systems fail when adversarial samples deviate from test conditions, leading to inconclusive benefits of such systems~\citep{carlini2017Adversarial,carlini2017MagNet, carlini2019Evaluating}.

\cite{athalye2018Obfuscated} showed that many of them appear robust only because they mask out the gradients, and proposed new types of attacks that circumvent gradient obfuscation.
In terms of detection systems, some studies have explored different statistical properties of adversarial samples for exploitation in detection~\citep{grosse2017Statistical}.
Recent studies have shifted their focus to the susceptibility of latent representations, positing them as the primary factor contributing to the adversarial vulnerability observed in deep neural networks. 
TRADES approach~\citep{zhang2019Theoretically} employs Kullback-Leibler divergence loss between a clean example and its adversarial counterpart to manipulate the decision boundary, aiming to achieve a more resilient latent space.

Research by~\cite{ilyas2019Adversarial} has demonstrated the presence of imperceptible features that aid in predicting clean examples but are susceptible to adversarial attacks. 
Conversely, rather than solely defending against adversarial attacks, ensuring robustness has emerged as a solution for model safety. 
The `randomized smoothing' technique, proposed empirically by~\cite{li2019Certified}, serves as a certified robustness measure. 
Subsequently,~\cite{cohen2019Certified} establishes the robustness guarantee of randomized smoothing against '2 norm adversarial attacks.'
Furthermore, efforts to enhance the efficacy of randomized smoothing are evident in the work by~\cite{salman2019Provably}, which involves a direct attack on the smoothed classifier.

\textbf{Adversarial Attacks and Adversarial Training for Object Detectors.}
Many effective attacks crafted for object detectors have been proposed recently. 
Most of them generate adversarial examples solely based on one individual loss (classification or localization loss) of the detection task \citep{xie2017Adversarial, chen2019ShapeShifter, song2018Physical, lu2017Adversarial}.
For instance, DAG \citep{xie2017Adversarial} optimizes over a loss function that misleads the detectors to produce incorrect classification results. 
Some other works \citep{li2019Exploring, liu2019DPatch} simultaneously attacks both the bounding box regression and classification to disable their predictions. 
To defend those attacks, \cite{zhang2019Adversarially, chen2021Robusta} extend adversarial training \citep{madry2018Deepa} to the scenario of object detection by leveraging the attacks sourced from both classification and localization domains. 

% Thus, adversarial distortions serve as type of worst-case analysis for network robustness.
% Its popularity has often led 'adversarial robustness' to become interchangeable with `robustness' in the literature (Bastani et al., 2016; Rauber et al., 2017).
% In the literature, new defenses (Lu et al., 2017; Papernot et al., 2017; Metzen et al., 2017; Hendrycks and Gimpel, 2017a) often quickly succumb to new attacks (Evtimov et al., 2017; Carlini and Wagner, 2017; 2016), with some exceptions for perturbations on small images (Schott et al., 2018; Madry et al., 2018).
% For some simple datasets, the existence of any classification error ensures the existence of adversarial perturbations of size O(d−1/2), d the input dimensionality (Gilmer et al., 2018b).
% For some simple models, adversarial robustness requires an increase in the training set size that is polynomial in d (Schmidt et al., 2018).

\subsection{Challenges in the evaluating natural corruptions, perturbations and distortions}

\textbf{Common Image Corruptions.}.
Various studies highlight the susceptibility of neural networks to common corruptions.
For instance, in a study by~\cite{hosseini2017Google}, impulse noise is employed to disrupt Google’s Cloud Vision API.
\cite{dodge2016Understanding} evaluated the performance of four cutting-edge image recognition models on out-of-distribution data, revealing that Convolutional Neural Networks are particularly sensitive to blur and Gaussian noise.
In another study,~\cite{dodge2017Study} utilized Gaussian noise and blur to highlight the superior robustness of human vision compared to neural networks, even after the networks underwent fine-tuning specifically for Gaussian noise or blur.

\cite{geirhos2018Generalisation} demonstrate that the performance of neural network declines more rapidly than human performance when faced with recognizing corrupted images, especially as the perturbation level increases across a diverse set of corruption types.
They find that fine-tuning on specific corruptions does not yield generalization, and the patterns of classification errors between network and human predictions differ.
Study by~\cite{azulay2019Why} delved into the lack of invariance of several state-of-the-art CNNs to small translations.
As such,~\cite{hendrycks2019Benchmarking} introduced a benchmark consisting of $19$ common corruptions to assess the robustness of recognition models against common corruptions to ease the understanding of effect of corruptions on a neural network.

\textbf{Robustness against Common Image Corruptions.}
To address the performance degradation on corrupted data, one approach involves preprocessing the data to eliminate the corruption.
\cite{mukherjee2018Visual} propose an approach to enhance  image quality of rainy and foggy images.
Similarly,~\cite{bahnsen2019Rain} propose algorithms to remove rain from images as a preprocessing step and report a subsequent increase in recognition rate.
However, a challenge with these approaches is their specificity to certain distortion types, limiting their generalization to other forms of distortions.

Another line of work seeks to enhance the classifier performance by the means of data augmentation, i.e. by directly including corrupted data into the training.
In an effort to reduce classifier fragility,~\cite{vasiljevic2017Examining} fine-tune on blurred images and suggest it is not enough to fine-tune on one type of blur to generalize to other blurs while fine-tuning on several blurs can marginally decrease performance.
Research by,~\cite{geirhos2018Generalisation} examine the generalization between different corruption types and find that fine-tuning on one corruption type does not enhance performance on other corruption types.
\cite{zheng2016Improving} also find that fine-tuning on noisy images can cause under-fitting, so they encourage the noisy image softmax distribution to match the clean image softmax.
\cite{dodge2017Quality} address under-fitting due to noisy images via a mixture of corruption-specific experts assuming corruptions are known beforehand.

In a different study,~\cite{geirhos2018ImageNettrained} train a recognition model on a stylized version of the ImageNet dataset~\citep{russakovsky2015ImageNet}, reporting increased general robustness against different corruptions as a result of a stronger bias towards ignoring textures and focusing on object shape.
\cite{hendrycks2019Benchmarking} present several methods that enhance performance on their corruption benchmark, including Histogram Equalization~\citep{delatorre2005Histogram, harvilla2012Histogrambased}, Multi-scale Networks~\citep{ke2017Multigrid, huang2018MultiScale}, Adversarial Logit Pairing~\citep{kannan2018Adversarial}, Feature Aggregating~\citep{xie2017Aggregated}, and Larger Networks.

\textbf{Robustness against Environmental Changes in Autonomous Driving}.
In recent years, weather conditions turned out to be a central limitation for state-of-the art autonomous driving systems \citep{sakaridis2018Semantic, dai2018Dark, chen2018Domain, lee2018Development}.
While many specific approaches like modelling weather conditions \citep{muller2015Robustness, sakaridis2018Model, sakaridis2018Semantic, bernuth2019Simulating, volk2019Robust} or collecting real \citep{wen2020UADETRAC,yu2020BDD100K, che2019City, caesar2020nuScenes} and artificial \citep{gaidon2016Virtual, ros2016SYNTHIA, richter2017Playing, johnson-roberson2017Driving} datasets with varying weather conditions, no general solution towards the problem has yet emerged.
\cite{radecki2016All} experimentally test the performance of various sensors and object recognition and classification models in adverse weather and lighting conditions.
\cite{vonbernuth2018Rendering} report a drop in the performance of a Recurrent Rolling Convolution network trained on the KITTI dataset \citep{geiger2012Are} when the camera images are modified by simulated raindrops on the windshield.
\cite{pei2022Practical} introduce VeriVis, a framework to evaluate the security and robustness of different object recognition models using real-world image corruptions such as brightness, contrast, rotations, smoothing, blurring and others.
\cite{machiraju2018Evaluation} propose a metric to evaluate the degradation of object detection performance of an autonomous vehicle in several adverse weather conditions evaluated on the Virtual KITTI dataset \citep{geiger2012Are}.
Building upon \cite{muller2015Robustness, volk2019Robust} study the fragility of an object detection model against rainy images, identify corner cases where the model fails and include images with synthetic rain variations into the training set.
They report enhanced performance on real rain images.
\cite{bernuth2019Simulating} model photo-realistic snow and fog conditions to augment real and virtual video streams.
They report a significant performance drop of an object detection model when evaluated on corrupted data.

\subsection{Self-Supervised Learning and Link to Robustness}

Recently, there has been a surge in the popularity of self-supervised learning \cite{gidaris2018Unsupervised, noroozi2016Unsupervised}, a technique that involves training a model on unlabeled data in a supervised manner by generating labels from the data itself.
The conventional self-supervised learning approach is to train the network to solve a manually defined (pretext) task for representation learning, which is subsequently applied to a specific supervised learning task such as image classification, object detection, or image segmentation.
One successful pretext task for self-supervised learning involves predicting the relative location of image patches \citep{noroozi2016Unsupervised, doersch2015Unsupervised, dosovitskiy2014Discriminative}. 
This method paved the way for the adoption of self-supervised learning.
\cite{gidaris2018Unsupervised} proposed learning image features by training deep networks to recognize 2D rotation angles, demonstrating significant improvement over previous self-supervised learning approaches.
Another effective strategy involves introducing corruption to images through processes like gray-scaling \citep{zhang2016Colorful} and random cropping \citep{pathak2016Context}, followed by restoring them to their original condition. 

Notably, preserving instance-level identity through contrastive learning has proven highly effective in generating rich representations for classification. Contrastive loss, either between two different views of the same images \citep{tian2020Contrastive} or two different transformed images from one identity \citep{chen2020Simple, he2020Momentum, tian2020What}, has shown comparable performance to fully-supervised models. 
These contrastive self-supervised learning frameworks aim to maximize the similarity of a sample to its augmentation while minimizing its similarity to other instances.
Recent studies indicate that leveraging unlabeled data contributes to the model's ability to acquire more robust representations \citep{carmon2019Unlabeled}. 
Additionally, \citep{hendrycks2019Using} demonstrates that a model trained with self-supervision exhibits improved robustness. 
The use of self-supervision signals, such as perceptual loss, has proven effective in de-noising adversarial perturbations, as seen in purifier network \cite{naseer2020Selfsupervised}. 
Even fine-tuning a pre-trained self-supervised learning model enhances robustness \citep{chen2020Adversarial}, and self-supervised adversarial training coupled with K-Nearest Neighbour classification enhances the robustness of KNN \cite{chen2020SelfSupervised}.

\section{Formal Mathematical Formulations}

Let us suppose that for the image classification problem, we consider a classifier 
$f_\theta : x \rightarrow y$ 
trained on samples from distribution $D$ such that 
$x \in \mathbb{R}^{m \times n \times c}$ 
be the  image which is to be classified.
Here $m, n$ is the width and the height of the image, and $c$ is the number of colour channels.

A neural network is composed of several neural layers linked together.
Each neural layer is composed of a set of perceptrons (artificial neurons).
Each perceptron maps a set of inputs to output values with an activation function.
Thus, function of the neural network (formed by a chain) can be defined as:
\begin{aequation}
f_\Theta(x) = g_{\theta_k}^{(k)}( \ldots g_{\theta_2}^{(2)}(g_{\theta_1}^{(1)}(x)))
\end{aequation}
where $g_{\theta_i}^{(i)}$ is the function of the $i^{\text{th}}$ layer of the network, and $ i = 1,2,3 \ldots k$ such that $k$ is the last layer of the neural network.
$\theta_i$ is the parameter of the $i^{\text{th}}$ layer which is optimised and consequently $\Theta = \{ \theta_1, \theta_2, \theta_3 \ldots \theta_k \}$ is set of all parameters of the neural network $f$ which is optimised.
In the image classification problem, 
$f(x) \in \mathbb{R}^{N}$ 
is the probabilities (confidence) for all the available $N$ classes.
Most classifiers are judged by their performance (often measured in accuracy) on test queries drawn from $D$, i.e.,
\begin{aequation}
\mathbb{P}_{(x, y) \sim D} ~~\underset{y}{\operatorname{argmax}} & & f(x) = y~~.
\end{aequation}

\subsection{Adversarial Perturbations}

Let us define adversarial samples $\hat{x}$ as:
\begin{aequation}
& \hat{x} = x + \epsilon_{x}
& \{ \hat{x} \in \R^{m \times n \times 3} \mid \underset{y}{\argmax} ~~ f(x) \ne \underset{\hat{y}}{\argmax}~~f(\hat{x})  \}
\end{aequation}
in which 
$\epsilon_{x} \in \R^{m \times n \times c}$ 
is the perturbation added to the input.
Here, $y$ and $\hat{y}$ are the respective classfications for $x$ and $\hat{x}$.

Making use of the definition of adversarial samples, adversarial robustness can be formally defined as the following optimization problem for untargeted black-box attacks:
\begin{aequation}
& \underset{\epsilon_{x}}{\text{minimize}} & & f(\hat{x})_y = f(x+\epsilon_{x})_y
& \text{subject to} & & \Vert \epsilon_{x} \Vert_p \leq \delta
\end{aequation}
Similarly optimization problem for the targeted black-box attacks can be defined as:
\begin{aequation}
& \underset{\epsilon_{x}}{\text{maximize}} & & f(\hat{x})_{\tilde{y}} = f(x+\epsilon_{x})_{\tilde{y}}
& \text{subject to} & & \Vert \epsilon_{x} \Vert_p \leq \delta
\end{aequation}
where $f()_c$ is the soft-label for the class $c$, and $y$ is the true class of sample $x$, whereas $\tilde{y}$ is target class for the the sample $x$. 
$p$ is the constraint on $\epsilon_{x}$ and $\delta$ is the threshold value for the constraint.
Thus, adversarial robustness can be formulated as,
\begin{aequation}
\underset{\Vert \epsilon_{x} \Vert_p ~\leq~ \delta}{\text{minimize}} & & \sP_{(x, y) \sim \train} ~~\underset{\hat{y}}{\argmax} & & f(\hat{x}) = y
\end{aequation}
Given such a dataset $\train$ and a model $f$, adversarial attacks aim towards finding the worst-case examples nearby by searching for the perturbation $ \epsilon $, which maximizes the loss.
We can define one such adversarial attack which tries to find perturbation within a certain radius from the sample (e.g., norm balls) as follows:
\begin{aequation}
\hat{x}^{i+1} = \Pi_{\Vert \epsilon_{x} \Vert_p ~\leq~ \delta}(\hat{x}^{i} + \alpha \cdot \text{sign}(\nabla_{\hat{x}^{i}} \Ls_{\text{CE}, \theta}(f(\hat{x}^{i}), y)))
\end{aequation}
where $\Vert \epsilon_{x} \Vert_p ~\leq~ th$ is the norm-ball around $x$ with radius $th$, and $\Pi$ is the projection function for norm-ball.
The $\alpha$ is the step-size of the attacks whereas sign(·) returns the sign of the vector and $\nabla_{p}(q)$ is the gradient of $p$ w.r.t $q$.
Further, $\mathcal{L}_{\text{CE}}$ is the cross-entropy loss for supervised training, and $i$ is the number of attack iterations.
This formulation generalizes across different types of gradient attacks.
For example, Projected Gradient Descent (PGD) \citep{madry2018Deepa} starts from a random point within the $x \pm th$ and perform $i$ gradient steps, to obtain the final adversarial sample $\hat{x}$.

\subsection{Common Corruptions}
We now consider a set of corruption functions $C$ such that $\mathbb{P}_C(c)$ approximate the real-world frequency of these corruptions.
We can now define, a classifier's robustness against corruptions as,
\begin{aequation}
\E_{c \sim C} [ \sP_{(x, y) \sim \train} ~~\underset{y}{\argmax} && f(c(x)) = y~~ ].
\end{aequation}
Thus, corruption robustness measures the classifier's average-case performance on classifier-agnostic corruptions $C$, while adversarial robustness measures the worst-case performance on small, additive, classifier-tailored perturbations.

\subsection{Adversarial Training}
The simplest and most straightforward way to defend against such adversarial attacks is to minimize the loss of adversarial examples, which is often called adversarial learning.
The adversarial learning framework proposed by \cite{madry2018Deepa} does solve the following non-convex outer minimization problem and non-convex inner maximization problem, as follows:
\begin{aequation}
\underset{\theta}{\argmin} & & \mathbb{E}_{(x, y) \sim D} [~~ \underset{\Vert \epsilon_{x} \Vert_p ~\leq~ th}{\text{maximize}} & & \Ls_{\text{CE}, \theta}(f(\hat{x}^{i}), y) ~~ ].
\end{aequation}
There are various adversarial learning frameworks, including PGD \citep{madry2018Deepa}, and TRADES \cite{zhang2019Theoretically} depending on how adversarial sample is optimized and how the classifier is optimized.

\subsection{Contrastive Self Supervised Learning}
The self-supervised contrastive learning framework aims to maximize the agreement between different augmentations of the same instance in the learned latent space while minimizing the agreement between different instances.
Let us define some notions, to project the image $x$ into a latent space, we use an encoder $f_\theta(·)$ network followed by a projector $g_\Pi(·)$, that projects the features encoded by the encoder into latent vector $z$.
Most self-supervised Contrastive learning methods uses a stochastic data augmentation $T$, randomly selected from the family of augmentations $\mathcal{T}$, including random cropping, random flip, random color distortion, and random grey scale.
Applying any two transformations, $T, \tilde{T} \sim \mathcal{T}$ , will yield two samples denoted $T(x)$ and $\tilde{T}(x)$, that are different in appearance but retains the instance-level identity of the sample.
We define $T(x)$’s positive set as $\{{x_{\text{pos}}}\} = \tilde{T}(x)$ from the same original sample $x$, while the negative set $\{x_{\text{neg}}\}$ as the set of pairs containing the other instances.
Then, the contrastive loss function using Noise Contrastive Estimation for Information (InfoNCE) \citep{oord2019Representation} $\mathcal{L}_{\text{InfoNCE}}$ can be defined as follows:
\begin{equation*}
\begin{aligned}
\mathcal{L}_{\text{InfoNCE}, \theta, \pi}(x) & = -\log \frac{ \sum_{\{z_{\text{pos}}\}} \exp(\text{sim}(z, \{z_{\text{pos}}\}) / \tau)}{\sum_{\{z_{\text{neg}}\}} \exp(\text{sim}(z, \{z_{\text{neg}}\}) / \tau)}
\end{aligned}
\end{equation*}
where $z$ is a latent vector obtained by the encoder $(f)$ and projector $(g)$ as $z = g_\Pi(f_\theta(x))$, whereas $\{z_{pos}\}$, and $\{z_{neg}\}$ are corresponding latent vectors of $\{{x_{\text{pos}}}\}$, and $\{x_{\text{neg}}\}$, respectively.
The $\text{sim}(u, v) = u^{T}v / \Vert u \Vert \Vert v \Vert$ denote cosine similarity between two vectors and $\tau$ is the temperature parameter.

\section{Methodology}

\subsection{Common Corruptions}

\begin{figure}[!t]
  \centering
  \includegraphics[width=\textwidth]{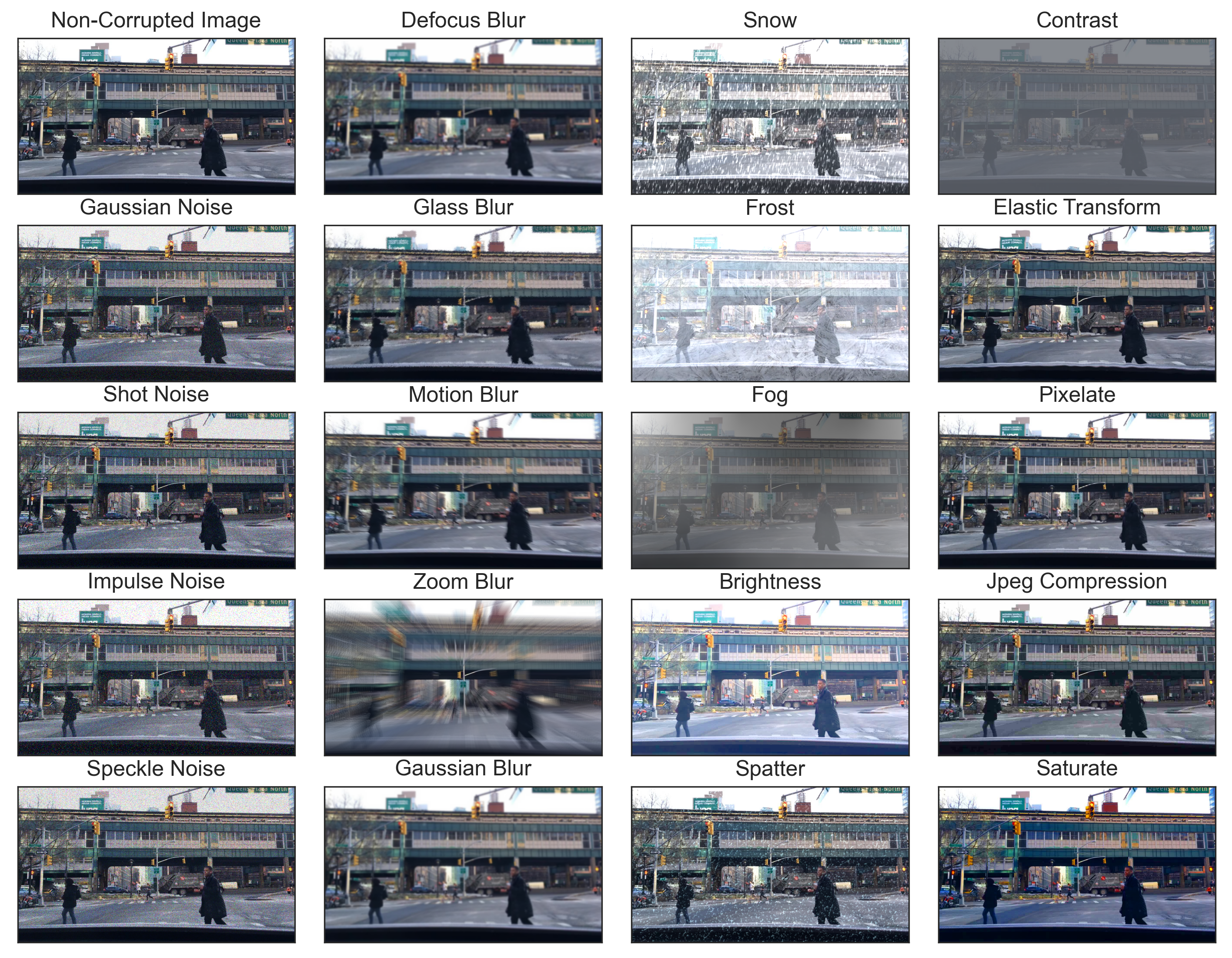}
  \caption{
    $19$ corruption types from \cite{hendrycks2019Benchmarking}, adapted to corrupt arbitrary images. 
    We corrupt a random image from BDD100k dataset \citep{yu2020BDD100K} with severity $3$ for all corruptions. 
    First column has Non-Corrupted Image and Noise Corruptions, Second column has Blur Corruptions, Third Column has Weather Corruptions, while the last column has Digital Corruptions. 
    Best viewed on screen.
  }
  \label{corruptions}
\end{figure}

The Common Corruptions benchmark \citep{hendrycks2019Benchmarking} introduced corrupted versions of commonly used classification datasets (ImageNet-C, CIFAR10-C) as standardized benchmarks which consists of $19$ diverse corruption types categoried into noise, blur, weather, and digital corruptions and each corruption type has five levels of severity.
This is because in real-world these corruptions can manifest themselves at varying intensities.
Since, the real-world corruptions also have variation even at a fixed intensity, to simulate these, the benchmark also introduces variation for each corruption when possible.
\Figref{corruptions} shows the $19$ types of corruption from \cite{hendrycks2019Benchmarking} applied on a random image from BDD100k dataset \citep{yu2020BDD100K}.
We  briefly describe each of the $19$ corruption in the benchmark below,

\begin{description}[itemsep=1.6pt, leftmargin=0pt]

  \item[Gaussian Noise,] this corruption can appear in low-lighting conditions.
  \item[Shot Noise,] also called Poisson noise, is electronic noise caused by the discrete nature of light itself.
  \item[Impulse Noise,] is a color analogue of salt-and-pepper noise and can be caused by bit errors.
  \item[Speckle Noise,] an additive noise where the noise added to a pixel tends to be larger if the original pixel intensity is larger.
  \item[Defocus Blur,] occurs when an image is out of focus.
  \item[Frosted Glass Blur,] appears with “frosted glass” windows or panels.
  \item[Motion Blur,] appears when a camera is moving quickly.
  \item[Zoom Blur,] occurs when a camera moves toward an object rapidly.
  \item[Gaussian Blur,] is a low-pass filter where a blurred pixel is a result of a weighted average of its neighbors, and farther pixels have decreasing weight in this average.
  \item[Snow,] is a visually obstructive form of precipitation.
  \item[Frost,] forms when lenses or windows are coated with ice crystals.
  \item[Fog,] shrouds objects and is rendered with the diamond-square algorithm.
  \item[Brightness,] varies with daylight intensity.
  \item[Spatter,] can occlude a lens in the form of rain or mud.
  \item[Contrast,] can be high or low depending on lighting conditions and the photographed object’s color.
  \item[Elastic Transformations,] stretch or contract small image regions.
  \item[Pixelation,] occurs when upsampling a lowresolution image.
  \item[JPEG Compression,] is a lossy image compression format which introduces compression artifacts.
  \item[Saturate,] is common in edited images where images are made more or less colorful.

\end{description}

\subsection{Evaluation Metric for Common Corruptions}

Common corruptions can be benign or destructive depending on their severity.
In order to comprehensively evaluate a classifier’s robustness to a given type of corruption, we score the classifier’s performance across five corruption severity levels and aggregate these scores.
The first evaluation step is to take a trained classifier $f$, and compute the performance on clean dataset ($P^{f}_{\text{clean}}$).
The second step is to test the classifier on each corruption type $c$ at each level of severity $s$ ($P^{f}_{c,s}$).

We then aggregate the classifier's performance for each of the $19$ corruption type $c$ as,
\begin{equation}
\begin{aligned}
\overline{P}^{f}_{c} = 1/5 \times \sum_{s=1}^{5} E_{f s,c}
\end{aligned}
\end{equation}
Finally, we aggreate the classifier's average performance on the common corruptions as,
\begin{equation}
\begin{aligned}
\overline{P}^{f} = 1/19 \times \sum_{c} E_{f s,c}
\end{aligned}
\end{equation}

\subsection{Adversarial Contrastive Learning}

\begin{figure}[!t]
  \centering
  \includegraphics[width=\textwidth]{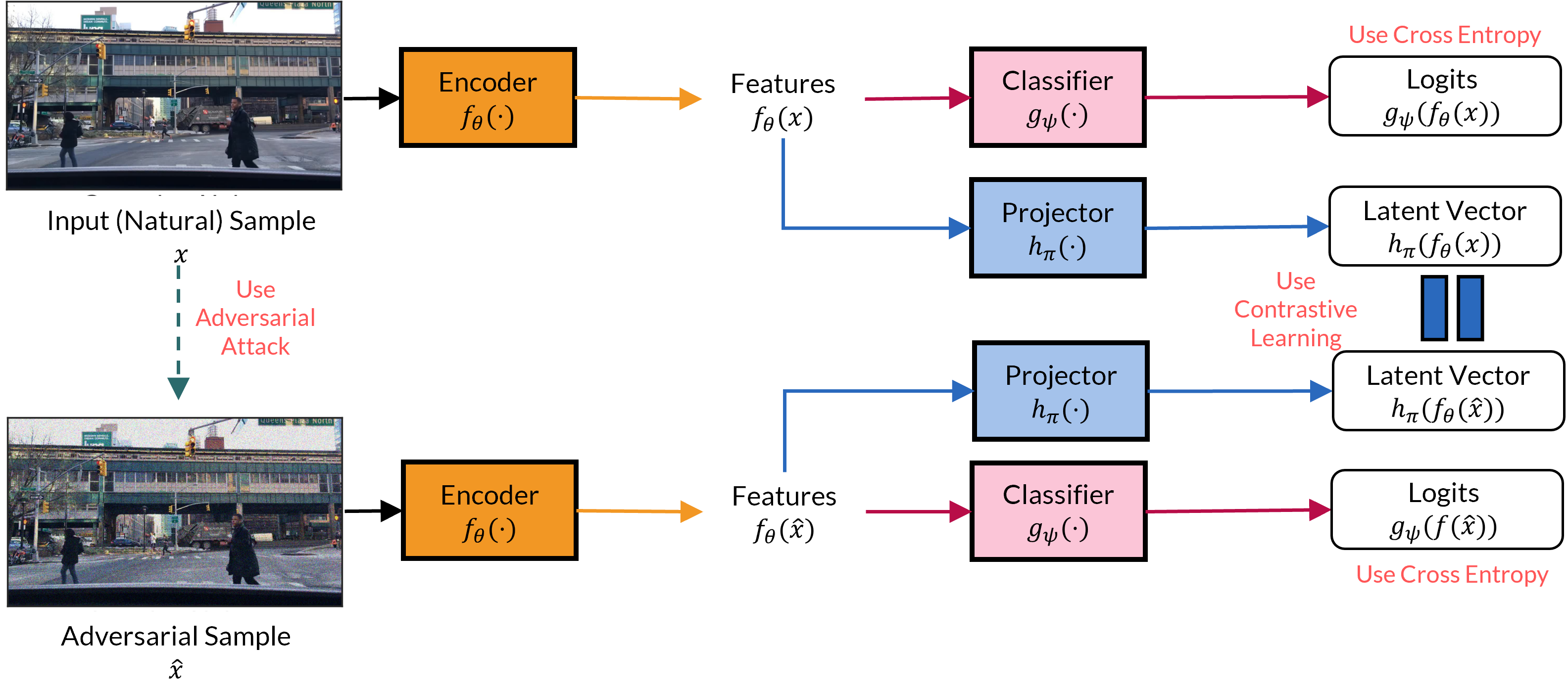}
  \caption{
    Illustration of the proposed Adversarial Contrastive Learning Approach.
  }
  \label{fig:acl}
\end{figure}

\begin{algorithm}[!t]
\caption{Adversarial Contrastive Learning}
\textbf{Input:} Dataset $D$, encoder $f$, encoder parameter $\theta$, classifier $g$, classifier parameter $\psi$,  projector $h$, projector parameter $\pi$
\begin{algorithmic}[1]
\For{each iteration $iter$ in the number of training iterations}

    \For{each $x$ in minibatch $B = \{x_1 \ldots x_m\}$}
        \State Generate adversarial examples from transformed inputs \Comment{Using Instance-Wise Attacks}
        \State $\hat{x}^{i+1} = \Pi_{\Vert \epsilon_{x} \Vert_p ~\leq~ th} \hat{x}^{i} + \alpha \cdot \text{sign}(\nabla_{\hat{x}^{i}} \mathcal{L}_{\text{CE}, \theta}(g(f(\hat{x}^{i})), y))$
    \EndFor
    \State $\mathcal{L}_{\text{Contrastive}} = \alpha \cdot \sum_{x=x_1}^{x_n}[ \mathcal{L}_{\text{InfoNCE},\theta, \pi}(h(f(x)), h(f(\hat{x}))) ] $ \Comment{Contrastive loss}
    \State $\mathcal{L}_{\text{Adversarial}} = \beta \cdot \frac{1}{2} \sum_{x=x_1}^{x_n}[ \mathcal{L}_{\text{CE},\theta, \psi}(g(f(x)), y) + \mathcal{L}_{\text{CE},\theta, \psi}(g(f(\hat{x})), y) ]$ \Comment{Adversarial loss}
    \State $\mathcal{L}_{\text{Total}} = \frac{1}{N} ~~ \mathcal{L}_{\text{Contrastive}} + \mathcal{L}_{\text{Adversarial}} $ \Comment{Total loss}
    \State Optimize the weights $\theta, \psi, \pi$ over $L_{\text{Total}}$
\EndFor
\end{algorithmic}
\label{algo:acl}
\end{algorithm}

Past methodologies involved the initial pretraining of the encoder $(f)$ and projector $(h)$ through contrastive learning, followed by the subsequent freezing of the encoder $(f)$ and finetuning of the classifier $(g)$. 
This procedure treats adversarial samples during the contrastive learning phase as minimally transformed samples, thereby not fully harnessing the true potential of adversarial examples to optimize the network. 
Consequently, since adversarial samples undergo minimal transformation, the benefits of contrastive learning are constrained. 
Additionally, studies by \cite{vargas2019Understanding, kotyan2021Deep} indicate that the impact of adversarial attacks initiates from the last layer of the evaluated network, and subsequent layers are affected in a butterfly effect manner.
Adversarial examples created in the contrastive phase using the encoder $(f_\theta)$ and projector $(h_\pi)$ might not be transferable to the final model consisting of 
encoder $(f_\theta)$ and finetuning of the classifier $(g_\psi)$ as, the parameters $\pi$ have changed to $\psi$.  
This puts a constraint on the robustness of final model using such methods depending on the adversarial transferability between the projector $(h_\pi)$ and classifier $(g_\psi)$.

In contrast, our proposal introduces a novel adversarial contrastive learning methodology that integrates adversarial training and contrastive learning simultaneously. 
\Figref{fig:acl} and Algorithm ~\ref{algo:acl} provide an illustration and the algorithm for the proposed adversarial contrastive learning. 
Initially, we generate the adversarial sample $\hat{x}$ from the given input image $x$. 
Subsequently, both the input image ($x$) and the adversarial sample ($\hat{x}$) are input into the encoder $f_\theta(\cdot)$ to obtain the features $f(x)$ and $f(\hat{x})$, respectively. 
These features are then fed into the classifier $g_\psi(\cdot)$ to acquire the logits $g(f(x))$ and $g(f(\hat{x}))$. 
The cross-entropy loss $\mathcal{L}_{\text{CE}}$ is computed using these logits, optimizing the parameters $\theta$ and $\psi$. 
Furthermore, the features are provided to the projector $h\pi(\cdot)$ to obtain the latent vectors $h(f(x))$ and $h(f(\hat{x}))$. 
The InfoNCE loss $\mathcal{L}_{\text{InfoNCE}}$ is computed using these vectors, optimizing the parameters $\theta$ and $\pi$. 
By utilizing the computed cross-entropy loss and InfoNCE loss together, the entire network is trained, and the parameters $\theta$, $\psi$, and $\pi$ of the encoder $(f_\theta)$, classifier  $(g_\psi)$, and projector $(h_\pi)$ are simultaneously optimized.

\section{Results}

\begin{table}[!t]
  \centering
  \caption{Performance of Faster RCNN with ResNet-50 as backbone with different training strategies for object detection and localisation task. Performance is measured in Mean Average Precision.}
  \resizebox{0.99\columnwidth}{!}{
  \begin{tabular}{l|l|rr}
    \toprule
    \multicolumn{2}{c|}{\textbf{Training}}& \textbf{Standard} & \textbf{Adversarial Contrastive Learning} \\
    \midrule
    \midrule
    \multicolumn{2}{l|}{\textbf{Natural Samples}} & 43.04\% & \x{47.94\%} \\
    \midrule
    \multirow{4}{*}{\textbf{Noise}}
    & \textbf{Gaussian Noise} & 25.83\% & \x{33.75\%} \\
    & \textbf{Shot Noise}     & 26.89\% & \x{33.94\%} \\
    & \textbf{Impulse Noise}  & 24.55\% & \x{31.14\%} \\
    & \textbf{Speckle Noise}  & 28.94\% & \x{33.24\%} \\
    \midrule
    \multirow{5}{*}{\textbf{Blur}}
    & \textbf{Defocus Blur}  & 27.87\% & \x{37.46\%} \\
    & \textbf{Glass Blur}    & 29.37\% & \x{34.77\%} \\
    & \textbf{Motion Blur}   & 21.93\% & \x{26.48\%} \\
    & \textbf{Zoom Blur}     & 26.71\% & \x{32.90\%} \\
    & \textbf{Gaussian Blur} & 31.13\% & \x{37.80\%} \\
    \midrule
    \multirow{5}{*}{\textbf{Weather}}
    & \textbf{Snow}       & \x{38.13\%} & 28.04\% \\
    & \textbf{Frost}      & 38.98\% & \x{46.41\%} \\
    & \textbf{Fog}        & \x{33.25\%} & 27.16\% \\
    & \textbf{Brightness} & 40.25\% & \x{46.12\%} \\
    & \textbf{Spatter}    & 32.94\% & \x{45.43\%} \\
    \midrule
    \multirow{5}{*}{\textbf{Digital}}
    & \textbf{Contrast}          & 32.08\% & \x{46.10\%} \\
    & \textbf{Elastic Transform} & 31.36\% & \x{39.05\%} \\
    & \textbf{Pixelate}          & 30.81\% & \x{34.39\%} \\
    & \textbf{JPEG Compression}  & 33.74\% & \x{41.35\%} \\
    & \textbf{Saturate}          & 36.35\% & \x{41.90\%} \\
    \midrule
    \rowcolor{Gray}
    \multicolumn{2}{c|}{\textbf{Average of 19 Corruptions}} & 31.11\% & \x{36.71\%} \\
    \bottomrule
  \end{tabular}
  }
  \label{tab:acl}
\end{table}

\textbf{Dataset.} 
BDD100k

\textbf{Model.} 
Faster-RCNN-ResNet-50

\textbf{Training Strategies} 
Standard Training and Adversarial Contrastive Learning

\textbf{Results.}
\Tableref{tab:acl} shows the result of object detector, Faster-RCNN-ResNet-50 trained using standard procedure and using adversarial contrastive learning recipe.
We can see also see that Adversarial Contrastive Learning on the standard BDD100k Dataset not only improves the performance by around $5\%$ of the Faster-RCNN on non-corrupted images but also improves robustness of Faster-RCNN in general by $5\%$ for common corruptions. 

We further notice that model trained with adversarial contrastive learning only degraded the performance for Snow and Fog. 
This might be because snow and fog corruptions are most distinct corruptions compared to perturbations applied by Adversarial Contrastive Learning. 
Further, snow and fog corruptions can easily occlude the small objects in the images making it impossible for the models to be detected. 
This results shows that Adversarial Contrastive Learning benefits improves the robustness of the model in general further suggesting that improving the latent representation of the neural networks can improves the better understanding of the features available in the image. 

\section{Conclusion}

In conclusion, our investigation into the impact of common corruptions on adversarial contrastive trained models has unveiled a notable advantage over models trained through the Standard Training Recipe. 
The observed heightened robustness of the former signifies a pivotal stride in bolstering the resilience of neural networks. 
Adversarial Contrastive Learning emerges as a potent methodology, refining latent representations to enhance a model's capacity to interpret intricate image features. 
This outcome extends beyond the immediate scope of our study, offering a promising avenue for fortifying object detectors in diverse real-world scenarios. 
By addressing the challenges posed by common corruptions, our findings underscore the broader implications of Adversarial Contrastive Learning, positioning it as a key player in advancing the reliability of artificial intelligence systems. 
As we anticipate that our work will stimulate further exploration and innovation, we believe that the integration of adversarial contrastive training stands to substantially contribute to the ongoing efforts in refining neural network representations and improving the overall resilience of machine learning models.

\bibliography{iclr2024_conference}
\bibliographystyle{iclr2024_conference}

%%%%%%%%%%%%%%%%%%%%%%%%%%%%%%%%%%%%%%%%%%%%%%%%%%%%%%%%%%%%

\end{document}

%% file: common.tex
\usepackage[T1]{fontenc}    % use 8-bit T1 fonts
\usepackage[utf8]{inputenc} % allow utf-8 input
\usepackage{algorithm, algorithmicx, algpseudocode}
\usepackage{amsmath, amsfonts, amssymb}
\usepackage{array}
\usepackage{booktabs}
\usepackage{collcell}
\usepackage{datatool}
\usepackage{enumitem}
\usepackage{environ}
\usepackage{etoolbox}
\usepackage{flushend}
\usepackage{graphicx}
\usepackage{lipsum}
\usepackage{microtype}
\usepackage{multirow}
\usepackage{nicefrac}
\usepackage{printlen}
\usepackage{soul}
\usepackage{svg}
\usepackage{url}
\usepackage{wrapfig}
\usepackage{xcolor,colortbl}
\usepackage{xstring}

\definecolor{Gray}{gray}{0.85}
\definecolor{cite}{HTML}{53769A}
\definecolor{ref}{HTML}{379030}
\definecolor{lightcornflowerblue}{rgb}{0.6, 0.81, 0.93}
\definecolor{lightkhaki}{rgb}{0.94, 0.9, 0.55}
\definecolor{lightmauve}{rgb}{0.86, 0.82, 1.0}
\definecolor{lightgreen}{rgb}{0.56, 0.93, 0.56}

\definecolor{PatternA}{RGB}{180, 22,  0   }
\definecolor{PatternC}{RGB}{23,  77,  127 }
\definecolor{PatternB}{RGB}{55,  144, 48  }

\usepackage[colorlinks=true, citecolor=cite, linkcolor=ref]{hyperref}

\newcolumntype{H}{>{\setbox0=\hbox\bgroup}c<{\egroup}@{}}

\newcommand{\com}[1]{\iffalse~#1~\fi}%
\newcommand{\x}[1]{{\underline{#1}}}%

\algrenewcommand{\algorithmiccomment}[1]{\hfill$\blacktriangleright$ #1}

\newcolumntype{X}{>{\columncolor{lightcornflowerblue}}c}
\newcolumntype{Y}{>{\columncolor{lightkhaki}}c}
\newcolumntype{Z}{>{\columncolor{lightmauve}}c}
\newcolumntype{P}{>{\columncolor{lightgreen}}c}

\newcommand{\noimage}{%
  \setlength{\fboxsep}{-\fboxrule}%
  \fbox{\phantom{\rule{100pt}{100pt}}File missing\phantom{\rule{100pt}{100pt}}}% Framed box
}
\let\includegraphicsoriginal\includegraphics%
\renewcommand{\includegraphics}[2][width=\textwidth]{\IfFileExists{#2}{\includegraphicsoriginal[#1]{#2}}{\noimage}}

\emergencystretch=\maxdimen%
\everypar{\looseness=-1 }

\newcounter{descriptcount}

\newcounter{CurrentRow}% = column of transposed table
\newcounter{CurrentColumn}
\newtoggle{DoneWithFirstRow}

\newcommand*{\FirstColumn}[1]{%
    \IfEq{\arabic{CurrentColumn}}{0}{%
        \global\togglefalse{DoneWithFirstRow}%
        \setcounter{CurrentRow}{1}% initial value
    }{%
        \global\toggletrue{DoneWithFirstRow}%
        \stepcounter{CurrentRow}%
    }%
    \setcounter{CurrentColumn}{0}%
    \NewData{#1}%
}
\newcommand*{\NewData}[1]{%
    \dtlexpandnewvalue%
    \stepcounter{CurrentColumn}%
    \iftoggle{DoneWithFirstRow}{%
        \dtlgetrow{TransposedTabularDB}{\arabic{CurrentColumn}}%
        \dtlappendentrytocurrentrow{\Alph{CurrentRow}}{#1}%
        \dtlrecombine%
    }{%
        \DTLnewrow{TransposedTabularDB}%
        \DTLnewdbentry{TransposedTabularDB}{\Alph{CurrentRow}}{#1}%
    }%
}%

\newcolumntype{+}{>{\global\let\currentrowstyle\relax}}
\newcolumntype{^}{>{\currentrowstyle}}
\newcolumntype{F}{>{\collectcell\FirstColumn}c<{\endcollectcell}}
\newcolumntype{C}{>{\collectcell\NewData}{c}<{\endcollectcell}}

\newtoggle{EncounteredDataRow}

\newsavebox{\TempBox}
\DTLnewdb{TransposedTabularDB}

\NewEnviron{Ttabular}[1]{%
    \setcounter{CurrentColumn}{0}%
    \global\togglefalse{EncounteredDataRow}%
    \savebox{\TempBox}{%
        \begin{tabular}{FCCCCCC}% over spaced tabular
            \BODY%
        \end{tabular}%
    }%
    \begin{tabular}{#1}\toprule%
    \DTLforeach*{TransposedTabularDB}{\Aa=A, \Ba=B, \Ca=C}{%
      \DTLiffirstrow{}{\\\midrule}%
        \Aa&\Ba&\Ca%
    }\\\bottomrule%
    \end{tabular}%
}%

\def\Tableref#1{Table~\ref{#1}}

\def\eqref#1{equation~\ref{#1}}

\usepackage{mathabx,graphicx}

\def\Figref#1{Figure~\ref{#1}}

\def\1{\bm{1}}
\newcommand{\train}{\mathcal{D}}

\newcommand{\E}{\mathbb{E}}
\newcommand{\Ls}{\mathcal{L}}
\newcommand{\R}{\mathbb{R}}

\DeclareMathOperator*{\argmax}{arg\,max}
\DeclareMathOperator*{\argmin}{arg\,min}

\newenvironment{aequation}
{\begin{equation*} \begin{aligned}}
{\end{aligned} \end{equation*}}

\DeclareMathAlphabet{\mathsfit}{\encodingdefault}{\sfdefault}{m}{sl}
\SetMathAlphabet{\mathsfit}{bold}{\encodingdefault}{\sfdefault}{bx}{n}

\def\sP{{\mathbb{P}}}